\documentclass[sigconf]{acmart}
\AtBeginDocument{%
  \providecommand\BibTeX{{%
    \normalfont B\kern-0.5em{\scshape i\kern-0.25em b}\kern-0.8em\TeX}}}

\setcopyright{acmcopyright}
\copyrightyear{2022}
\acmYear{2022}
\acmDOI{10.1145/1122445.1122456}

\usepackage{graphicx}
\graphicspath{ {./images/} }
\usepackage{subcaption}
\usepackage{amsmath}
\usepackage[flushleft]{threeparttable}
\usepackage{xspace}
\usepackage{makecell}
\usepackage{adjustbox}
\usepackage{array}
\usepackage{pict2e}
\usepackage{makecell}
\newtheorem{problem}{Problem}
\newcommand{\pcdetector}{\textsc{ColdGuess}}
\DeclareMathOperator*{\argmax}{argmax}
\newcommand{\method}{\textsc{ColdGuess}\xspace}

\newcommand{\lgbmgraph}{\textsc{ColdGuess}-Naive\xspace}
\newcommand{\lgbm}{lgbm \xspace}
\newcommand{\lightgbm}{lightgbm}
\newcommand{\consolidated}{consolidated nodes}
\newcommand{\homoinf}{homogeneous influence}
\newcommand{\homoinfps}{homogeneous influence\xspace}

\newcommand{\WIS}{Type1 \xspace}
\newcommand{\USN}{Type2 \xspace}
\newcommand{\Damaged}{Type3 \xspace}
\newcommand{\OtherNPE}{Type4 \xspace}
\newcommand{\Defective}{Type5 \xspace}
\newcommand{\VM}{Type6 \xspace}
\newcommand{\CCR}{Type7 \xspace}
\newcommand{\EPCR}{Type8 \xspace}
\newcommand{\ASIN}{product \xspace}
\newcommand{\ASINs}{products \xspace}
\newcommand{\ASINg}{Product \xspace}
\usepackage{multirow,float}

\usepackage{xcolor}
\definecolor{lb}{RGB}{44, 139, 183}

\usepackage{slashbox}
\usepackage{diagbox}
\begin{document}

%%
%% The "title" command has an optional parameter,
%% allowing the author to define a "short title" to be used in page headers.
\title{\method: A General and Effective Relational Graph Convolutional Network to Tackle Cold Start Cases }

%%
%% The "author" command and its associated commands are used to define
%% the authors and their affiliations.
%% Of note is the shared affiliation of the first two authors, and the
%% "authornote" and "authornotemark" commands
%% used to denote shared contribution to the research.
\author{Bo He}
\email{bhe@amazon.com}
\affiliation{%
  \institution{Amazon Inc.}
  \country{US}
}

\author{Xiang Song}
\email{xiangsx@amazon.com}
\affiliation{%
  \institution{AWS AI Education and Research}
  \country{US}
}

\author{Vincent Gao}
\email{vincegao@amazon.com}
\affiliation{%
  \institution{Amazon Inc.}
  \country{US}
}

\author{Christos Faloutsos}
\authornote{On leave from CMU}
\email{faloutso@amazon.com}
\affiliation{%
  \institution{AWS AI Education and Research}
  \country{US}
}

%%
%% By default, the full list of authors will be used in the page
%% headers. Often, this list is too long, and will overlap
%% other information printed in the page headers. This command allows
%% the author to define a more concise list
%% of authors' names for this purpose.

%%
%% The abstract is a short summary of the work to be presented in the
%% article.
\begin{abstract}
% Each year Amazon receives hundreds of millions of product complaints globally. These result in sub-optimal buying experience and erode customer trust.
% Each year Amazon receives 350MM+ product complaints globally. These erode customer trust and result in over \$4.1B losses as measured by downstream impact.
% To reduce product related issues, we developed \method, a product complaint catcher which proactively identifies risky seller/ASIN/offer listings through a graph neural network.
% Traditional machine learning models like logistic regression, support vector machine, and boosting trees are widely used in risk detection. However, they face two limitations:
Low-quality listings and bad actor behavior in online retail websites threatens e-commerce business as these result in sub-optimal buying experience and erode customer trust. When a new listing is created, how to tell it has good-quality? Is the method effective, fast, and scalable? Previous approaches often have three limitations/challenges: (1) unable to handle cold start problems where new sellers/listings lack sufficient selling histories. (2) inability of scoring hundreds of millions of listings at scale, or compromise performance for scalability. (3) has space challenges from large-scale graph with giant e-commerce business size. To overcome these limitations/challenges, we proposed \pcdetector, an inductive graph-based risk predictor built upon a heterogeneous seller-\ASIN graph, which effectively identifies risky seller/\ASIN/listings at scale. %through a graph neural network.
% Within the graph, seller and seller nodes are connected through their associated bank account, credit card, emails, etc., and seller and ASIN nodes are connected if a seller offers that ASIN.
\method tackles the large-scale graph by \textit{\consolidated}, and addresses the cold start problems using \textit{\homoinf}\footnote{See definition of \consolidated and \homoinf in Section 4.3}.
% It allows us to model and propagate risks across sellers and listings, and achieve better prediction than traditional models in the cold start cases.
The evaluation on real data demonstrates that \method has stable performance as the number of unknown features increases. It outperforms the \lightgbm\footnote{lightgbm is abbreviated as lgbm in the paper.} by \textit{up to 34 pcp} ROC AUC in a cold start case when a new seller sells a new \ASIN. The resulting system, \pcdetector, is effective, adaptable to changing risky seller behavior, and is already in production. %at Amazon.

% \method augments the product verification and risky seller prevention system and has led to significant precision improvement in bin checks.
% This large-scale graph based complaint detector will \textit{be in production in early July} of 2021, making daily predictions on \textit{55MM+} listings with a goal of 5\% YoY reduction in customer complaints.
\end{abstract}

%%
%% Keywords. The author(s) should pick words that accurately describe
%% the work being presented. Separate the keywords with commas.

%%
%% This command processes the author and affiliation and title
%% information and builds the first part of the formatted document.
\settopmatter{printfolios=true}
\maketitle

\section{Introduction}
% Each year Amazon receives hundreds of millions of product related complaints globally. These degrade the customer shopping experience and erode customer trust.
% This also has a big negative impact on the Amazon brand.
% These complaints are categorized into eight types, such as Expired, Used Sold as New, Defective, Damaged, Wrong Item Sent, Version Mismatch and Other Negative Product Experience. 

Detecting low-quality listings and bad actor behavior is an important task for e-commerce websites ~\cite{chen2015big, hui2016reputation}. When a new product arrives, how to tell it is good-quality? When a bad actor creates a new account, how to detect it before any fraudulent behavior endangers customers shopping experience. Is the detection mechanism fast and scalable? Can it score millions of sellers and hundreds of millions of listings? When a new seller is associated with multiple existing seller accounts through either weak or strong relations, can this method tell which relation to emphasize? These are the research problems we focus on in this work.

\begin{figure}[t]
    \centering
    \begin{subfigure}[b]{1.0\columnwidth}
        \centering
        \includegraphics[width=0.6\linewidth]{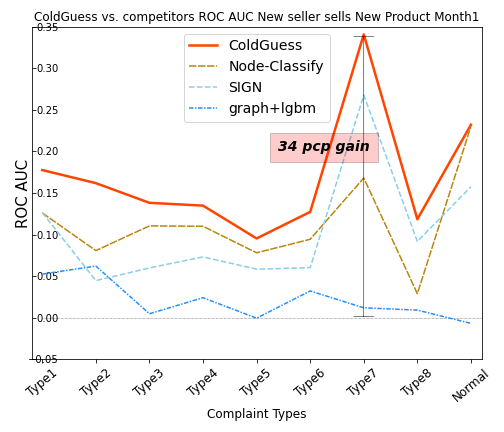}
        \caption{\method wins all benchmark models in the cold start case when new sellers sell new \ASINs.}
        \label{fig:sub1}
    \end{subfigure}
    \begin{subfigure}[b]{1.0\columnwidth}
        \centering
      \includegraphics[width=0.6\linewidth]{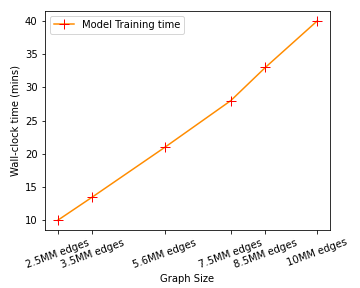}
      \caption{\method scales linearly: model training time vs graph size.}
      \label{fig_train}
    \end{subfigure}
\caption{\method's performance and scalability.}
\label{fig:intro}
\end{figure}

Traditional machine learning models like
logistic regression, support vector machine\cite{kim2003constructing, mareeswari2016prevention}, and boosting trees~\cite{ke2017lgbm} are widely used in risk detection. However, they often face following limitation/challenges: 

\begin{itemize}
    \item unable to model new sellers/offers (\textit{the Cold Start Problem}). New sellers/offers lack sufficient histories for traditional models to identify as risky. As large e-commerce companies continuously expand to new countries, the cold start problem becomes more common and critical.
    \item inability of scoring hundreds of millions of listings at scale, or compromise performance for scalability. 
    \item non-graph based methods (e.g. logistic regression, boosting tree, etc.) are not able to leverage seller-seller and seller-\ASIN  \footnote{For example, Seller Smith sells Nike shoes. Here Smith is a seller, Nike shoes is a  product. The Nike shoes sold by Smith is an offer listing. In this paper, we use offer, listings, and offer listings interchangeably.}linkage information\footnote{e.g., sellers are associated if they share the same information or similar characteristics.}. Linkage information has proven to be valuable as risky sellers/offers are often found clustered.\footnote{For example, majority of sellers who share the same information with a risky seller are also found risky.} Graph-based method face the challenges of memory space when applied to large scale of graph with giant e-commerce business size. 
\end{itemize} 

To overcome these limitations, we proposed \pcdetector, a graph neural network based risk detector which leverages the seller-seller and seller-\ASIN linkage information to identify risky seller/\ASIN/listings at scale. It handles the large-scale graph by \textit{\consolidated}\footnote{See definition in Section 4.3}, and addresses the cold start problem by \textit{\homoinf}\footnote{See definition in Section 4.3} and message passing from existing sellers/\ASINs to new sellers/\ASINs. As shown in Table ~\ref{tab:ds}, there are many competitors, but none of them has all the features that \method offers.

We evaluated \method and compared it with four competitors on a full spectrum of sample data and three simulated cold start scenarios (i.e. new offers, new sellers, and new sellers who sell new \ASIN). 
% We compared its performance with lgbm\cite{ke2017lgbm}, 
% %\lgbmgraph, an algorithm combining one-hop graph propagation with lgbm, 
% SIGN\cite{frasca2020sign} and RGCN\cite{schlichtkrull2018modeling}
We found that \method wins with increasing margins as the severity of cold start cases increases. As shown in Figure~\ref{fig:intro}, \method outperforms \lgbm by up to \textbf{34 percentage points (pcp) in AUC} in the extreme cold start scenario when a new seller sells a new \ASIN. %It has 20\% to 10 times relative improvement in average precision \footnote{measured by PC AUC} across different complaint types. 

%\begin{figure}[t]
%\centering
%  \includegraphics[width=0.75\linewidth]{Figures/2020-11_seller_asin.png}
%  \label{fig:sub1}
%\caption{\method wins all benchmark models in the cold start case when new sellers sell new \ASINs.}
%\label{fig:intro}
%\end{figure}

The main contributions and advantages of \method are:

% \begin{enumerate}
%     \item Previous GNN models in CTPS mainly focused on homogeneous graphs/features, and are sub-optimal in the heterogeneous setting. We proposed PC-Detector framework allows risk propagation across heterogeneous graph in a highly multi-relational data set. 
%     \item In the state-of-art RGCN, missing values on edge features cannot be inferred from their neighbors since aggregation and updates only happen on node features. To deal with the missing values on edge features, We augmented an offers' edge features by concatenating it with its neighbors' edge features. An edge's neighbor edge is defined as edges connected to that offer's head and tail nodes. 
%     \item We found RGCN has superb performance and outperforms LGBM in cold start cases, though it cannot beat LGBM when evaluated on overall data. 
% \end{enumerate}
\begin{itemize}
\item \textbf{General:} The inductive \method framework can handle (a) \textit{Dynamic heterogeneous graphs built on highly relational data} with tens of millions of nodes, hundreds of millions of edges, and massive node and edge features.
%both node and edge features on the graph. 
%Previous GNN models in production was either transductive or mainly focused on homogeneous graphs without features, and are sub-optimal in the heterogenous setting. \method framework works for heterogeneous graph constructed from a highly multi-relational data set and allows node feature propagation over different node types and edge types. %Note from xiangsx: This statement is not valid, thus I comment it out. 
% \item \textit{Cold start problem}. \method achieves stable performance with increasing number of missing features on new offer/seller/ASIN scenarios. It outperforms LGBM by up to \textbf{30\%} in an extreme cold start case where a seller sells a new ASIN.
(b) \textit{Missing values in edge feature.} In previous approaches to graph modeling, missing edge features cannot be inferred from their neighbors since propagation only happens on the nodes. To deal with it, \method augmented edge features by concatenating it with its neighboring edges.
% by concatenating an edge feature and a summary of its neighbors' edge features.

\item \textbf{Effective}, especially for \textit{Cold Start Problem}: \method achieves stable performance with increasing number of missing features in cold start scenarios. It outperforms \lgbm by up to \textit{34 pcp} AUC in the extreme cold start case where a new seller sells a new \ASIN. 
%improves its performance from basic RGCN setting by concatenating edge feature with its neighbor's edge features. In the state-of-art RGCN, missing values on edge features cannot be inferred from their neighbors since aggregation and updates only happen on node features. To deal with the missing values on edge features, we augmented an offers' edge features by concatenating it with its neighbors' edge features. An edge's neighbor edge is defined as edges connected to that offer's head and tail nodes.
%\item \textbf{Effective:} \method has superb performance and outperforms lgbm in cold start cases. 
\item \textbf{Fast and Scalable:} 
\method can scale to a business size graph with hundreds of millions of edges in inference. It takes only 2 minutes to make 1MM+ predictions. The training on a graph with 10MM+ edges takes 40 minutes on a p3.8xlarge EC2 instance equipped with 4 V100 GPU and 224GB memory. \method scales linearly with the input size in both training and inference, as shown in Figure~\ref{fig:scalability}.
\end{itemize}

% \newcommand{\myrotate}[1]{\rotatebox{80}{#1}}
% \begin{table}[t]
%   \centering
%   \caption{\method matches all specs, while competitors miss one or more of the features.}
%   \begin{tabular}{l|cccc||c}
%     \hline
%   \diagbox{Property}{Method}
% % \backslashbox{Property}{Method}
%     &\myrotate{lgbm~\cite{ke2017lgbm}}&\myrotate{graph+lgbm}&\myrotate{SIGN\cite{frasca2020sign}}
%     &\myrotate{RGCN\cite{schlichtkrull2018modeling}}&\myrotate{\method}\\
%     %   Method & lgbm & graph+lgbm & SIGN & RGCN & \method \\    
%       \hline
%       node connectivity   &  & $\surd$ & $\surd$ & $\surd$ & $\surd$\\

%       heterogeneous information &  &         &         & $\surd$ & $\surd$ \\
%       edge connectivity   &  &         &         &         & $\surd$ \\
%      \hline   
%   \end{tabular}
%   \label{tab:ds}
% \end{table}

\begin{table}[t]
  \centering
  \caption{\method matches all specs, while competitors miss one or more of the features.}
  \begin{tabular}{|l|ccc||cc|}
  \hline
  \diaghead{\theadfont heterogeneous information more sp}{~~Property}{Method}
    &\thead{\rotatebox{70}{\lgbm~\cite{ke2017lgbm}}}&\thead{\rotatebox{70}{SIGN\cite{frasca2020sign}}}
    &\thead{\rotatebox{70}{RGCN\cite{schlichtkrull2018modeling}}}&\thead{\rotatebox{70}{\lgbmgraph}}&\thead{\rotatebox{70}{\method}} \\  
      \hline
      node connectivity         &  & $\surd$ & $\surd$ & $\surd$ & $\surd$\\

      heterogeneous information &  &         &   $\surd$ &  & $\surd$ \\
      edge connectivity   &  &         &         &         & $\surd$ \\
      \hline
  \end{tabular}
  \label{tab:ds}
\end{table}

\section{Preliminaries}
\textbf{Notations}. The notations used in the following sections are listed in Table~\ref{tbl:notation}.

\begin{table*}[t]
\begin{threeparttable}
%\scriptsize
\caption{Notation and descriptions.}
\label{tbl:notation}
\centering
\begin{tabular}{ll}
\toprule
$\mathcal{G}$ & A graph. \\
$\mathcal{V}$ & The set of nodes. \\
$\mathcal{E}$ & The set of edges. \\
$\mathcal{R}$ & The set of relations. \\
$\mathcal{X}$ & The node feature space where $x_i$ represents the node feature of $v_i \in \mathcal{V}$. \\
$\mathcal{Y}$ & The edge feature space where $y_i$ represents the edge feature of $e_i \in \mathcal{E}$. \\
$\mathcal{H}$ & The node embedding matrix where $h_v$ represents the embedding vector of $v$. \\
$\mathcal{Z}$ & The defect type label space where $z_i$ represents the $i$-th label vector. \\
$\mathcal{S}$ & The set of nodes $v \in \mathcal{V}$ with the node type of Seller. \\
$\mathcal{O}$ & The set of edges $e \in \mathcal{E}$ with the edge type of (Seller, Offer, \ASIN). \\
$\mathcal{P}$ & The set of nodes $v \in \mathcal{V}$ with the node type of \ASIN. \\
%$s$ & A Seller node that $s \in \mathcal{S}$. \\
%$o$ & An Offer edge that $o \in \mathcal{O}$. \\
%$a$ & An \ASIN node that $a \in \mathcal{A}$. \\
$l$ & An offer listing $\left<s, o, p\right>$ with $s \in \mathcal{S}$, $o \in \mathcal{O}$ and $p \in \mathcal{P}$. \\
$L$ & A set of offer listings. \\
$\mathbf{s}, \mathbf{p}$, $\mathbf{o}$ & The feature vectors of seller node $s$, \ASIN node $p$ and offer edge $o$.\\
\bottomrule
\end{tabular}
\end{threeparttable}
\end{table*}

\paragraph{Graph} A graph is composed of nodes and edges with relations defining the heterogeneity $\mathcal{G} = (\mathcal{V}, \mathcal{E}, \mathcal{R})$, where $\mathcal{V}$ is the set of nodes, $\mathcal{E}$ is the set of edges, $\mathcal{R}$ is the set of relations. Each edge $e \in \mathcal{E}$ belongs to only one relation $r \in \mathcal{R}$.
Let $\mathcal{X}$ be the node feature space such that $\mathcal{X}=(x_1, \dots, x_N)^T$ where $x_i$ represents the node feature of $v_i$. Let $\mathcal{Y}$ be the edge feature space such that $\mathcal{Y}=(y_1, \dots, y_M)^T$ where $y_i$ represents the edge feature of $e_i$. Let $\mathcal{Z}$ be the label space.

\paragraph{Graph Convolutional Network}
%In recent years, graph neural networks(GNN) ~\cite{ying2018graph,scarselli2008graph,hu2020open,yu2021self} have shown great success in graph based machine learning tasks. 
GNNs~\cite{ying2018graph,scarselli2008graph} are a series of multi-layer feedforward neural networks that propagate and transform layer-wise features. 
% % Among these models, a graph convolutional network(GCN)~\cite{kipf2016semi} architecture is widely employed, which relies on the layer-wise message passing scheme. 
% %Formally, the $(l+1)$-th layer of a GCN\footnote{The vanilla GCN model as well as its varieties are design for homogeneous graphs which has no relations.} is defined as:
% %\begin{equation} \label{eq:1}
% %    h^{(l+1)} = \sigma(D^{-\frac{1}{2}}AD^{-\frac{1}{2}}h^{(l)}W^{(l)})
% %\end{equation}
% %where $W^{(l)}$ denotes a trainable weight matrix of the $l$-th layer, $\sigma((·)$ is an activation function, and $h^{(l)}$ represents the $l$-th layer node representation. We initialize $h_v^0=x_v$. 
% %The representation vectors of the nodes in the last layer are usually feed into downstream tasks such as node classification or edge classification.
Among these models, relational graph convolutional network (RGCN)~\cite{schlichtkrull2018modeling} was designed to operate on large-scale heterogeneous graphs. The (l+1)-th layer node representation of $v$ in RGCN is defined as:
 \begin{equation} \label{eq:2}
      h_v^{(l+1)} = \sigma(\sum_{r \in R}\sum_{u \in \mathcal{N}_v^r}\frac{1}{c_{v,r}}W_r^{(l)}h_u^{(l)}+W_0^{(l)}h_v^{(l)})
 \end{equation}
\noindent where $\mathcal{N}_v^r$ denotes the set of neighbors of node $v$ under relation $r \in R$, $h_u^{(l)}$ represents the $l$-th layer node representations of nodes $u \in \mathcal{N}_v^r$, $h_v^{(l)}$ represents the $l$-th layer node representation of $v$ itself,
$W_r^{(l)}$ denotes a trainable weight matrix of the $l$-th layer corresponding to relation $r$ and $W_0^{(l)}$ denotes a trainable weight matrix of the $l$-th layer corresponding to $v$ itself. $c_{v,r}$ is a problem-specific normalization value that is usually defined as $c_{v,r}=|\mathcal{N}_v^r|$, $\sigma$ denotes the activation function.
The representation vectors of the nodes in the last layer are usually fed into downstream tasks such as node classification or edge classification.

\section{Related Work}
\paragraph{Cold Start}
%DropoutNet: Addressing Cold Start in Recommender Systems
Cold start problem gains increasing interest in last few years, and was widely researched in recommendation systems.~\cite{lam2008addressing,zhou2011functional,lika2014facing, bi2020heterogeneous} The challenge of cold-start recommendation comes from sparse user-item interactions for new users or items. Most recent works~\cite{hu2018leveraging, wang2019knowledge, lu2020meta} leverage heterogeneous graph to capture richer semantics via higher-order graph structures and meta information. 
Hu, Binbin et al.~\cite{hu2018leveraging} showed that side information such as meta-path based context is useful to alleviate the cold start problem.
Lu, Yuanfu et al.~\cite{lu2020meta} leveraged meta-learning to address cold-start recommendation.
However, unlike user-item recommendation scenarios where users do not directly connect with other users, there are rich linkage information between sellers in a seller-\ASIN graph in e-commerce. Furthermore, unlike edges between users and items which contain little information~\cite{harper2015movielens, asghar2016yelp}, the listing edges in a seller-\ASIN graph usually contain rich information on its selling histories. Our proposed
\method leverages such linkage information to address cold start problem. 
Other than only using GNN, \method uses \textit{\homoinf} to further boost the performance and \textit{\consolidated} to handle the scalability issue.

%DropoutNet~\cite{volkovs2017dropoutnet} shows that neural network models can be explicitly trained for cold start through random input dropout. It claims that cold start is equivalent to the missing data problem where preference information is missing. However, it does not show whether input dropout can work with graph data. 

% fraud detection
\paragraph{Fraud Detection}
Fraud detection is essential for e-commerce~\cite{zhou2018state}. Industries use machine learning methods to catch fraudulent transactions, such as SVM~\cite{kim2003constructing, mareeswari2016prevention}, neural network~\cite{barse2003synthesizing, fu2016credit}, and boosting tree~\cite{sahin2013cost,rushin2017horse}. 
However, these methods can not leverage the linkage information of heterogeneous network of modern web graphs. Hence, GNN-based fraud detection attracts increasing attention. FdGars~\cite{wang2019fdgars}, DCI~\cite{wang2021decoupling} and SemiGNN~\cite{wang2019semi} tend to summarize the abnormal patterns automatically using GNNs to detect fraudsters. These methods focus on detecting abnormal nodes in a graph, while \method focuses on detecting abnormal edges. 
LIFE~\cite{li2021live} and GAS~\cite{li2019spam} incorporate both node and edge features during message passing for fraudulent transactions detection. However, we found that incorporating GNN with \textit{\homoinf} is more effective than merely using GNN in risky offer listing detection. 

\section{Method}
% In this section, we first define the product complaint detection problem and related cold start scenarios. Then we show how a Seller-ASIN graph is constructed. Finally, we present \pcdetector, which leverages the seller-seller and seller-ASIN linkage information to address the cold start problem. 

\subsection{Problem Definition}
We use customer complaints as labels for low-quality listings. There are eight types of complaints. We formulate this complaint detection as a multi-label classification problem:

%Given an offer listing $l$ with its associated 
%Semi-formal Problem:
\begin{problem}
Given (1) labelled listings $\mathcal{L}$ in 1 out of $k$ categories, %$k=8$
(2) the numerical and categorical features for sellers, \ASINs, and offers denoted as $\mathbf{s}, \mathbf{p}$, $\mathbf{o}$ respectively, and (3) Seller-\ASINg graph $\mathcal{G}$, find a classifier $f$ which maximizes the likelihood of offer listing $l$ having complaint type $z$:
    \begin{equation}
    \hat{z} = \argmax f(z|s, p, o, \mathcal{G})
    \end{equation}
\label{pro:pro}
\end{problem}
%\textbf{Given}
%\begin{itemize}
%    \item labelled offer listings in 1 out of $k$ categories. $k=8$
%    \item numerical and categorical features for sellers, \ASINs, and offers
%    \item Seller-\ASINg graph $\mathcal{G}$
%\end{itemize}

%\textbf{Find} 
%\begin{itemize}
%    \item a classifier $f$ which maximizes the likelihood of offer listing $l$ will have complaint type $z$:
%    \begin{equation}
%    \hat{z} = \argmax f(z|s, p, o, \mathcal{G})
%    \end{equation}
%\end{itemize}

In addition to Problem~\ref{pro:pro}, the classifier should be efficient to handle \textit{large-scale graphs} and work effectively with \textit{cold start problems}.
% With all information available, traditional models such as boosting tree and logistic regression works well in risk detection. However, when it is a new offer or a new seller, traditional models usually fail due to lack of sufficient history. We call this \textit{Cold Start Problem} in which limited information is available for new offers/sellers/ASINs. Amazon faces three kinds of cold start problems as illustrated in Figure~\ref{fig:eval}: 
% Risk detection models often face cold start problem on new sellers/\ASINs lacking sufficient history for models to identify as risky.
We summarize the cold start problems into three cases:
%as illustrated in Figure~\ref{fig:eval} and below:
% However, when an new offer is proposed by a seller or a new seller comes to make offer listings, the performance of traditional ML models will downgrade as there will be limited information associated with certain new offers or sellers. We call this the \textit{cold start problems}. In general, there are three kinds of cold start problems:
\begin{itemize}
    \item \textbf{New offer case}: An existing seller creates a new offer under an existing \ASIN. Both seller and \ASIN nodes of that new offer exist in the Seller-\ASIN graph. The offer edge is newly added to the graph, and has limited history information. 
    % In this case, the new offer will have no selling history and thus all its edge features will be set as zeros except features known once the offer is created, such as listing price. %listing price is known as long as the offer is created.
    \item \textbf{New seller case}: An new seller creates a new offer under an existing \ASIN. Only the \ASIN node of the new offer exists in the Seller-\ASINg graph. The seller node and offer edge are newly added, and have limited seller and offer information.
    % In this case, the new seller will have no selling record that the corresponding node features are set as zeros. However, the new seller can still be connected with other seller nodes in the graph through relations like bank account, credit card and etc. Similar to the \textit{New offer} case, the new offer edge have limited features.
    \item \textbf{New seller sells a new \ASIN case}: A new seller creates a new \ASIN which was never sold on Amazon. Neither of the seller nor \ASIN nodes are related to any existing listings on the Seller-\ASIN graph. Therefore, there is limited information of the seller, \ASIN, and offer. 
    
    % In this case, as the new ASIN has no ASIN level selling history that the ASIN node features are set with zeros except GL product group. 
    %An ASIN's GL group would be known even it is a new ASIN. 
    % Like the \textit{New seller case}, both the seller node and the offer edge have limited features.
\end{itemize}
In the last two cases, new sellers can be connected with existing sellers if they possess associated information. The severity of cold start increases from new offer case to new seller case, and then to new seller sells a new \ASIN case.

\begin{figure}
\centering
  \includegraphics[scale=0.29]{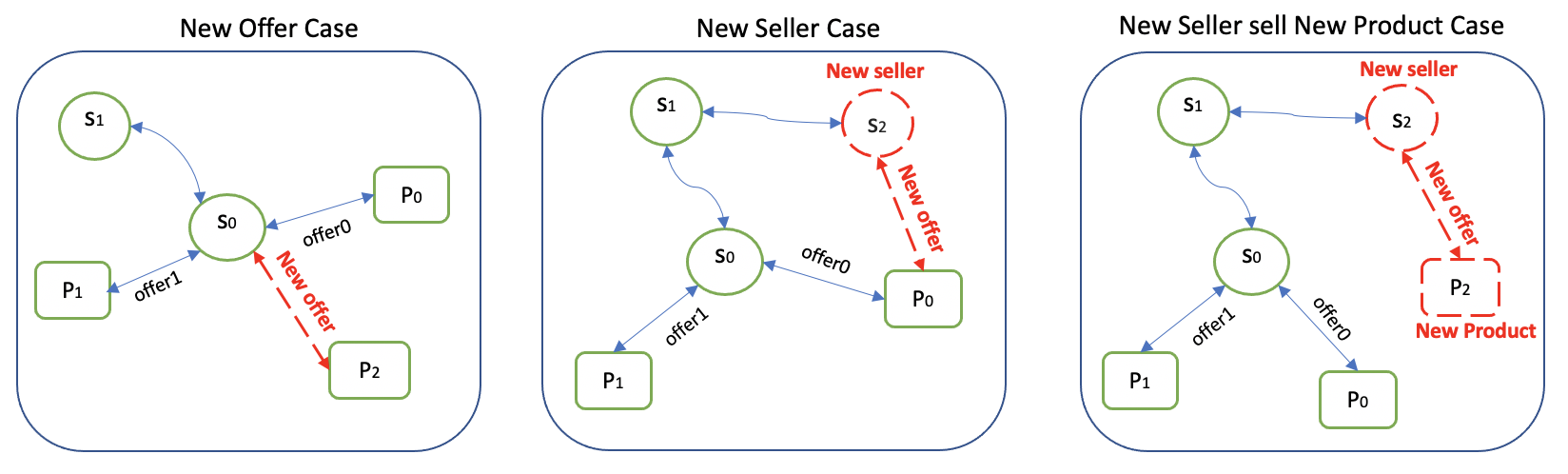}
\caption{Cold start scenarios: new offer case (left), new seller case (middle), new seller sell new \ASIN case (right)}
\label{fig:eval}
\end{figure}

\subsection{Seller-\ASINg Graph} \label{graph-cons}
We constructed Seller-\ASINg graph as a heterogeneous graph with seller and \ASIN nodes.
% Within the graph, sellers are connected if they are associated via bank account, credit card, phone, verified phone, international bank account, email, cis\_sign\_in, siv\_name\_poc, tims\_tin\_id\footnote{Those relations are chosen based on Risk Relation team's SOP. It states that a seller can be determined as risky if she is connected with a risky seller through any two or three combination of above eight relations}.
Within the graph, sellers are connected if they possess associated information.
%\footnote{Those relations are chosen based on Risk Relation team's SOP. It states that a seller can be determined as risky if she is connected with a risky seller through any two or three combination of above eight relations}.
A seller is connected with an \ASIN if the seller sells that \ASIN. 
The resulting graph $\mathcal{G}$ has two node types and nine relations (i.e. eight seller-seller relations plus a seller-\ASIN relation (i.e. offer edge)), as well as massive features on seller nodes, \ASIN nodes, and offer edges. 
The \ASIN and seller level features are encoded as the node feature vectors. The offer level features are encoded as edges feature vectors. \ASINg node features include product type, \ASIN's selling history, etc. Seller node features include seller's orders, selling history, etc. Offer edge features include list price, orders, offer's selling history, etc. 
% ASIN node features include ASIN age, glance views, GL product group, ASIN's selling history, defect count and ratios, etc. Seller node features include ratings, geographic data, seller's selling history, defect count and ratios, etc. Offer edge features include list price, competitor's information, offer's selling history, defect count and ratios, etc. 

We call this graph design as \textit{\consolidated} because we treat offers as edges instead of nodes. It reduces the number of edges by half as compared to using offers as nodes. 

\begin{figure}
\centering
  \includegraphics[scale=0.24]{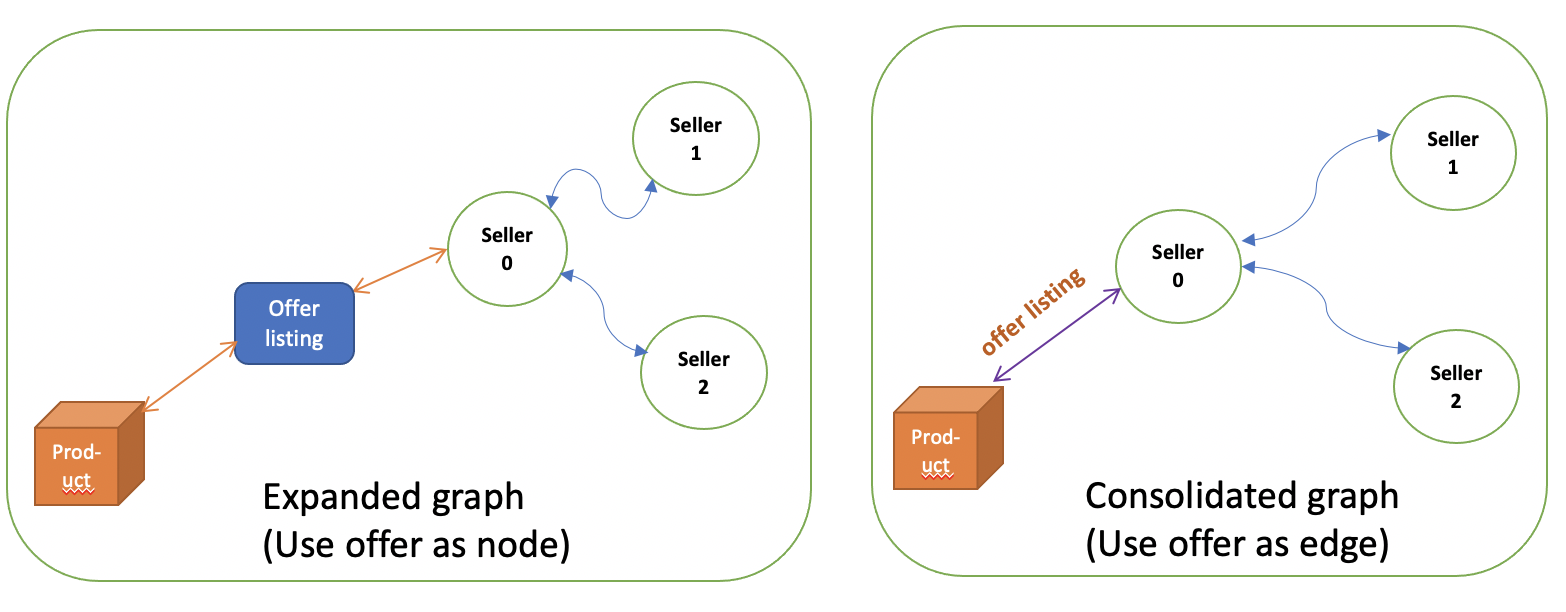}
\caption{A toy example of an expanded graph with offers as nodes (left) vs. a consolidated graph with offers as edges (right)}
\label{fig:offer_node}
\end{figure}

\paragraph{Why the 'consolidated node' proposal?} Alternatively, we can construct the graph by using offers as nodes (Figure ~\ref{fig:offer_node} right). An offer node connects a seller and a \ASIN nodes. To predict the complaint type of an offer can leverage the RGCN node classification framework. We evaluated this alternative set up on 4 test sets over 4 months' dataset, one for each month, and found: 1) \method consistently outperforms in at least 7 out of 9 classes when being evaluated on the full spectrum of data and cold start cases. It is superior to RGCN benchmark with increasing margins as the severity of cold-start cases grows.%, as shown in Figure ~\ref{fig:202011} and ~\ref{fig:202012}.
2) \method saves GPU memory space by 65\% and computation time by 15\% when both methods are evaluated on a 4-GPU P3.8xlarge EC2. The \textit{\consolidated} design reduces the graph size without losing offer information as \method extracted offer information via an edge embedder\footnote{See Section 4.3 Module 2: Edge Embedder}. 

% In certain scenarios, a consolidated graph (Figure ~\ref{fig:offer_node} right) propagates risk better than a fully expanded graph in which edges are added as node (Figure ~\ref{fig:offer_node} left). For example, in a consolidated graph a seller's risk can directly propagate to the connecting seller. However, in a fully expanded graph a risky seller will be exonerated by connecting to a bank with high credit score. 

\subsection{Proposed Method: \pcdetector}
% We proposed \pcdetector, which leverages the seller-seller and seller-ASIN linkage information and addresses the cold start problems that usually fail traditional machine learning models.
% Given the seller, ASIN, and offer features, and the Seller-ASIN graph $\mathcal{G}$, \pcdetector predicts the probability of offer listing $l$ to have complaint type $z$:
% \begin{equation}
%     P(z|l,\mathcal{G}) = f_{\pcdetector}(l, \mathcal{G}, \mathcal{X}, \mathcal{Y})
% \end{equation}

\begin{figure}
\centering
  \includegraphics[scale=0.30]{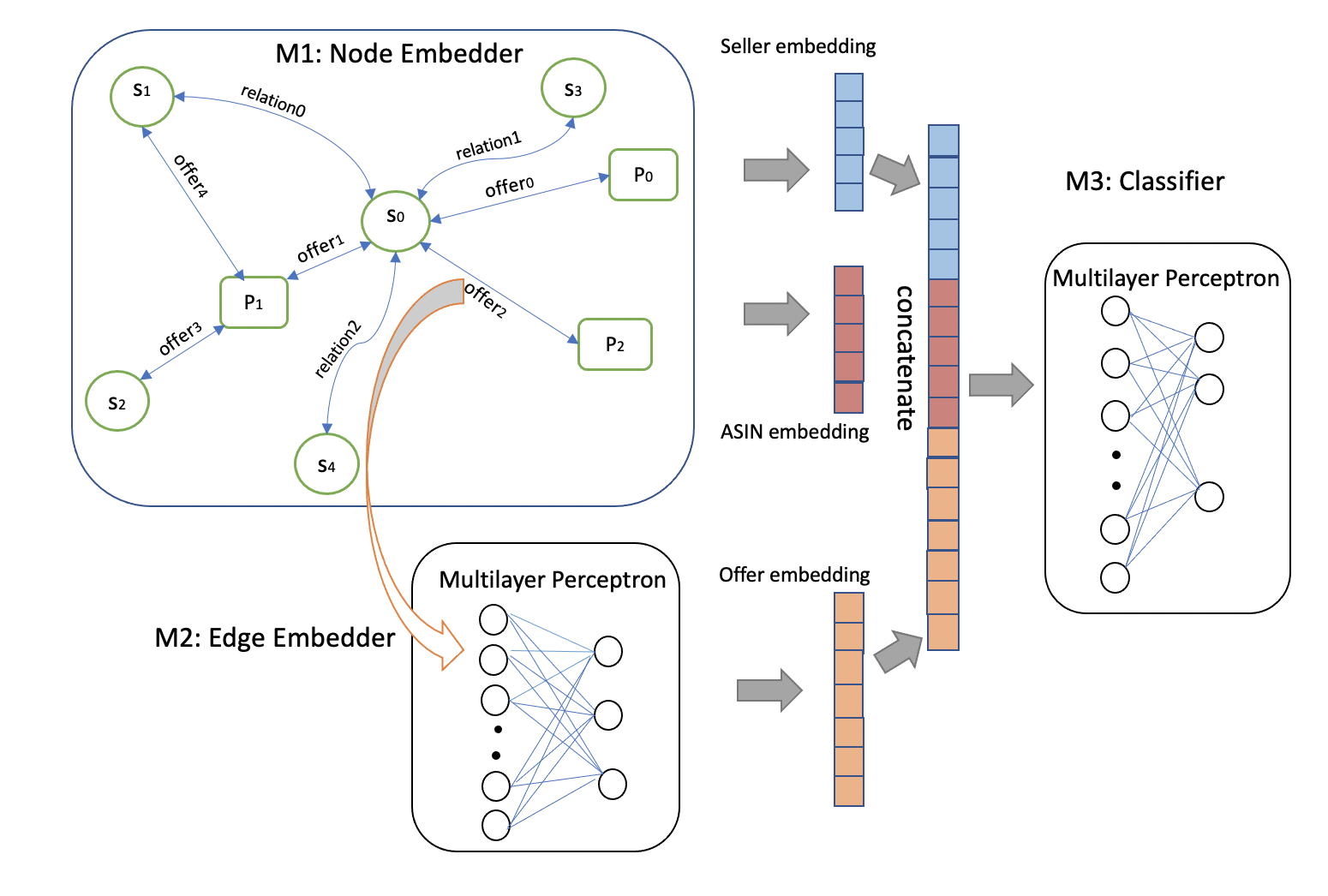}
\caption{The overall architecture of \pcdetector}
\label{fig:pcd}
\end{figure}

\method contains three modules: 
1) A node embedder built on a 3-layer RGCN which encodes raw seller and \ASIN node features into two embedding vectors $emb_s$ and $emb_p$.
2) An edge embedder built on a 2 layer multi-linear perceptron (MLP) which encodes offer edge features $\mathbf{o}$ into an embedding vector $emb_o$
3) A final MLP classifier which concatenates the seller, \ASIN, and offer embedding vectors ($emb_s \| emb_p \| emb_o$) and predicts the product complaint type. The overall model architecture is shown in Figure~\ref{fig:pcd}.  
% of $l$ based on the concatenated embedding of $emb_s$, $meb_a$ and $emb_o$.

% \pcdetector is implemented using DGL~\cite{wang2019deep} and PyTorch~\cite{paszke2019pytorch}. In \method, a node embedder encoded seller and ASIN node features into 64-dim embedding vectors using a 3-layer relational GCN. An edge embedder encoded offer edge features into 105-dim embedding vectors using a 2-layer MLP. A final 2-layer MLP classifier combines the seller, ASIN, offer embedding and predicts the product complaint type. Both \pcdetector and LGBM models were trained on data of 2020/10 and evaluated on the data from 2021/11 to 2021/04. 

% a multi-layer RGCN model which aggregates the multi-hop neighbor information of both the seller node $s$ and the ASIN node $a$ of an offer listing $l$ and outputs two low dimensional embedding vectors $emb_s$ and $meb_a$ for $s$ and $a$ respectively, 2)a Multi-Linear Perceptron (MLP) model which projects the initial feature vector $\mathbf{o}$ of the offer edge $o$ into a low dimensional embedding vector $emb_o$, and 3) a Multi-Linear Perceptron (MLP) model which predicts the product complaint type of $l$ according to $emb_s$, $meb_a$ and $emb_o$.

% \pcdetector can aggregate useful information from neighbors of both ASIN and seller nodes through multi-layer message passing. This is proven to be beneficial in solving the cold start problem when the new sellers and ASINs usually have limited features.

% We examine each of the three modules in details as well as their implementation below:
\paragraph{Module 1: Node Embedder} It creates the seller and \ASIN embedding ($emb_s$ and $emb_p$) using RGCN~\cite{schlichtkrull2018modeling}. Since seller and \ASIN node features have different dimensions and meanings, we projected them into the same embedding space before feeding them into the graph convolution. 

%The (l+1)-th layer node representation of $v$ in RGCN is defined as:
%\begin{equation} \label{eq:2}
%    h_v^{(l+1)} = \sigma(\sum_{r \in R}\sum_{u \in \mathcal{N}_v^r}\frac{1}{c_{v,r}}W_r^{(l)}h_u^{(l)}+W_0^{(l)}h_v^{(l)})
%\end{equation}
%where $\mathcal{N}_v^r$ denotes the set of neighbors of node $v$ under relation $r \in R$, $h_u^{(l)}$ represents the $l$-th layer node representations of nodes $u \in \mathcal{N}_v^r$, $h_v^{(l)}$ represents the $l$-th layer node representation of $v$ itself,
%$W_r^{(l)}$ denotes a trainable weight matrix of the $l$-th layer corresponding to relation $r$ and $W_0^{(l)}$ denotes a trainable weight matrix of the $l$-th layer corresponding to $v$ itself. $c_{v,r}$ is a problem-specific normalization value that is usually defined as $c_{v,r}=|\mathcal{N}_v^r|$.

% As our Seller-\ASIN graph usually has millions of nodes and edges, it is infeasible to feed the entire graph into GPU during training. 
Due to the large scale of Seller-\ASINg graph, we take a graph-based mini-batch approach~\cite{hamilton2017inductive} for propagation. We sampled a batch of 1024 offer edges and extracted the corresponding ego network which covers the neighbors $3$-hops away from the source edges, and fed this mini-batch into the $3$-layer RGCN module.

% In each mini-batch, we randomly sample $N$ offer listings $L$. Then we extract a $k$-hop ego network from $\mathcal{G}$ for each seller node $s$ and ASIN node $a$ appearing in $l \in L$, where $k$ is the number of RGCN layers. A graph $G^\prime$ is created by compacting all these ego networks. 
% Finally, the graph $G^\prime$ and the offer listings $L$ are feed into the RGCN model as:
% \begin{equation}
%     (\mathcal{H}_s, \mathcal{H}_a) = g_{RGCN}(L, G^\prime)
% \end{equation}
% where $g_{RGCN}$ is the RGCN model. The output is the node embedding matrices $\mathcal{H}_s$ and $\mathcal{H}_a$ for the seller nodes and ASIN nodes in $L$ respectively.

\paragraph{Module 2: Edge Embedder} It creates offer edge embedding ($emb_o$) via a MLP model. It takes an offer's edge features and its neighboring edge features as input. 
%  takes the initial feature vector $\mathbf{o}$ of an offer edge $o$ and produces the low dimensional embedding vector $emb_o$. 
Offer edges usually have missing values, especially in cold start cases. We enrich the offer feature vector $\mathbf{o}_{o}$ by concatenating it and its neighbors. The neighboring edge is defined as edges connected to the offer’s head and tail node (i.e. \ASIN and seller nodes). We create an offer feature summary for both the \ASIN node and the seller node as:
\begin{equation}
    \begin{aligned}
        \mathbf{o}_{p} = \psi(\mathbf{o}_1, \dots, \mathbf{o}_n), \text{where } o_i \in \mathcal{N}_p \\
        \mathbf{o}_{s} = \psi(\mathbf{o}_1, \dots, \mathbf{o}_n), \text{where } o_i \in \mathcal{N}_s
    \end{aligned}
\end{equation}
where $\psi$ is a aggregation function
%We tested mean, min-pool and max-pool, and arrived at similar performance. }
such as mean, max-pool, etc.~\footnote{We use mean in implementation.}, $o_i$ are offer features of edge $i$, $\mathcal{N}_p$ and $\mathcal{N}_s$ are the sets of an offer's neighboring edges connected to its \ASIN and seller nodes respectively, $\mathbf{o}_{p}$ and $\mathbf{o}_{s}$ are the summary of feature vectors of offers connected to \ASIN and seller nodes. The final offer feature vector is defined as $\mathbf{o^\prime} = \mathbf{o}_{o} \| \mathbf{o}_{p} \| \mathbf{o}_{s}$, where $\|$ denotes the concatenation operation. Eventually, $\mathbf{o^\prime}$ is fed into the MLP classifier to generate the offer embedding $emb_o$.
%This method of adding neighbors' edge embedding was proved effective as it improves the AUC by about 1-2\% in new offer cases. 

The concatenation of neighboring offer information with target offer is a direct offer to offer influence. We call it \textit{\homoinf}. It is in contrast to the non-\homoinfps from offer to offer when using offers as nodes, where information from an offer node will first pass to a seller/\ASIN node and then to another offer node.

\paragraph{Why the '\homoinf?} \textit{\homoinf} more effectively keeps offer information as compared to the 2-hop propagation (offer -> seller/\ASIN -> offer) when offers are used as nodes. Some offer features contain strong signals for low-quality listings. However, they will be averaged out by its neighbors during propagation in non-\homoinfps setting. The reduced signal becomes more difficult for model to detect. Moreover, the message passing process will blend intermediate seller/\ASIN node information and added noises signals. In the \textit{\homoinf} method, offer features do not go through the propagation process and are largely kept. %The evaluation in Section 5 also demonstrates \method with \textit{\homoinf} outperforms the RGCN node classification (i.e.non-\homoinf) in at least 7 out of 9 classes in full spectrum of data as well as cold start cases. As shown in Figure ~\ref{fig:202011} and ~\ref{fig:202012}, this \textit{\homoinf} method wins with increasing margins when cold start cases becomes more severe. 

% \paragraph{MLP model for complaint type prediction}
\paragraph{Module 3: Final Classifier}
This classifier first concatenates $emb_s$, $emb_p$ and $emb_o$ as inputs and predict the probability of the offer listing $l$ that will have a defect type $z$ via a multi-layer perceptron. 
\begin{equation}
    emb_l = f(emb_s \| emb_p \| emb_o)
\end{equation}
where $\|$ denotes the concatenation operation and $f$ denotes the MLP layer. Finally it uses sigmoid to get the prediction result.
%\begin{equation}
%    p = \frac{\exp{(emb_{l_i})}}{\sum_{j=1}^{d}\exp{(emb_{l_j})}}, %\text{where } i \in [1, \dots, d]
%\end{equation}
%where d is the dimension of $emb_l$ and $p$ is the probability of $l$ belonging to category $z$

\paragraph{Train Objective}
We train \pcdetector in an end to end fashion using mini-batches of offer listings $L$. We define the cross-entropy loss as the classification error:
\begin{equation}
    \mathcal{L} = - \frac{1}{|L|}\sum_{l \in L}z_l\log{p_l} + (1-z_l)\log{(1-p_l)}
\end{equation}
where $z_l \in \{0, 1\}$ is the ground-truth for the offer listing $l$ and $p_l$ is the output of the \pcdetector. For a fair comparison with \lgbm, which trained 9 binary models one for each class, in this paper we trained \method in the same fashion. In production, we used the sum of 9 cross entropy loss as the objective function and trained \method as one model to predict all classes. It reduces the inference time without much performance loss except one class lowered by less than 2\% absolute AUC.

We optimize the whole model using stochastic gradient descent. The loss is back-propagated over the entire framework to update all the parameters.

\section{Evaluation}
\subsection{Evaluation Setup}
% We conducted extensive experiments on real world datasets to evaluate \pcdetector. 
% We compared the performance of \pcdetector with lgbm, a boosting tree based learning algorithms which takes seller, \ASIN, offer features as input and outputs a score for each defect class.

We conducted extensive evaluation on \method using real data, and compared it with three baseline models. For a fair comparison, all models are trained using the same feature set.
\begin{itemize}
\item{\textbf{\lgbm}} We chose \lgbm~\cite{ke2017lgbm} as benchmark because (1) it often outperforms other machine learning methods in modeling tabular data~\cite{sergei2021boost}. (2) it can show the value of using graph information. 
% trained the lgbm model on offer features and its associated seller and \ASIN features. and the corresponding defect labels as input. The input feature of an offer listing $\left<s, o, a\right>$ is defined as $x_s\|y_o\|x_a$, where $\|$ denotes the concatenation operation.

\item{\textbf{SIGN}} SIGN~\cite{frasca2020sign} is a graph deep learning architecture which sidesteps the need for graph sampling by using graph convolutional filters of different sizes that are amenable to efficient precomputation. It allows extremely fast training and inference. We trained SIGN using the same graph as training \pcdetector. We chose SIGN as benchmark to evaluate the value of using learnable weights in \pcdetector. 

\item{\textbf{RGCN}} RGCN~\cite{schlichtkrull2018modeling} is a graph neural network designed for modeling relational graph data. We chose RGCN as benchmark to compare the performance between using offers as edges vs. as nodes. 

\end{itemize}

We also developed a simplified version of \pcdetector:
\begin{itemize}
\item{\textbf{\lgbmgraph}} \lgbmgraph combines one-hop graph propagation with \lgbm. It first fills the missing seller features using one-hop graph propagation, and then feeds the seller features along with \ASIN and offer features to a \lgbm model. A seller's missing features is filled by simply averaging her connected sellers, with all seller-seller relation types equally weighted. We do not fill offer/\ASIN missing features using neighboring offers/\ASINs because two \ASINs could be totally different things, and test results shows it under-performs filling missing seller features only. 
%Two offers connecting with the same \ASIN usually comes from different sellers and their features will diverge.
\end{itemize}

%\subsubsection{Experimental Setting}
We used the implementation of \lgbm from LightGBM project~\footnote{https://github.com/microsoft/LightGBM}.
We used the implementations of SIGN and RGCN from DGL~\cite{wang2019deep} and implemented \method using DGL and PyTorch~\cite{paszke2019pytorch}. We used Adam optimizer to train our model. 
For SIGN and \pcdetector, we chose a 3-layer structure. For RGCN, we chose a 6-layer structure to collect comparable neighborhood information as \pcdetector. The embedding dimension of the hidden layers is set to 64.

We chose ROC AUC as the major evaluation metric because (1) rank-based metrics are robust; (2) it shows the FPR/TPR for all possible threshold values and (3) unlike PR AUC which varies as the label distribution of underlying data changes, ROC AUC is easier for comparison across data sets with different label distributions. All numbers reported in the following sections are presented using percentage point (pcp) in AUC, notated as \%.

% \begin{figure*}[t]
% \centering
%     \includegraphics[width=.25\textwidth]{Figures/2020-11_All Data.png}\hfill
%     \includegraphics[width=.25\textwidth]{Figures/2020-11_offer.png}\hfill
%     \includegraphics[width=.25\textwidth]{Figures/2020-11_seller.png}\hfill
%     \includegraphics[width=.25\textwidth]{Figures/2020-11_seller_asin.png}\hfill
%     \caption{Performance of \method, lgbm, \lgbmgraph, SIGN and RGCN on a full spectrum of sample data, new offer case, new seller case, and new seller sell new \ASIN case using data of Month1. The severity of cold start increases from left to right.}
%     \label{fig:202011}
% \end{figure*}

\begin{figure*}[t]
\centering
    \includegraphics[width=0.95\textwidth]{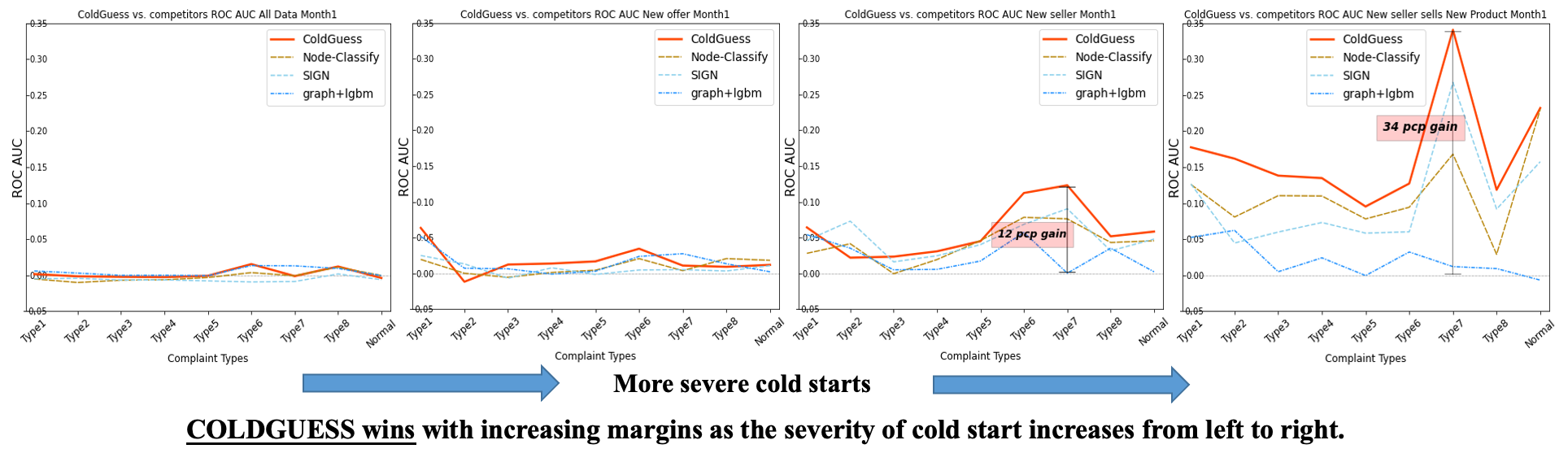}\hfill
    \caption{Performance of \method, \lgbm, \lgbmgraph, SIGN and RGCN on a full spectrum of sample data, new offer case, new seller case, and new seller sell new \ASIN case using data of Month1.}
    \label{fig:202011}
\end{figure*}

% \begin{figure*}[t]
% \centering
%     \includegraphics[width=.25\textwidth]{Figures/2020-12_All Data.png}\hfill
%     \includegraphics[width=.25\textwidth]{Figures/2020-12_offer.png}\hfill
%     \includegraphics[width=.25\textwidth]{Figures/2020-12_seller.png}\hfill
%     \includegraphics[width=.25\textwidth]{Figures/2020-12_seller_asin.png}\hfill
%     \caption{Performance of \method, lgbm, \lgbmgraph, SIGN and RGCN on a full spectrum of sample data, new offer case, new seller case, and new seller sell new \ASIN case using data of Month2. The severity of cold start increases from left to right.}
%     \label{fig:202012}
% \end{figure*}

\begin{figure*}[t]
\centering
    \includegraphics[width=0.95\textwidth]{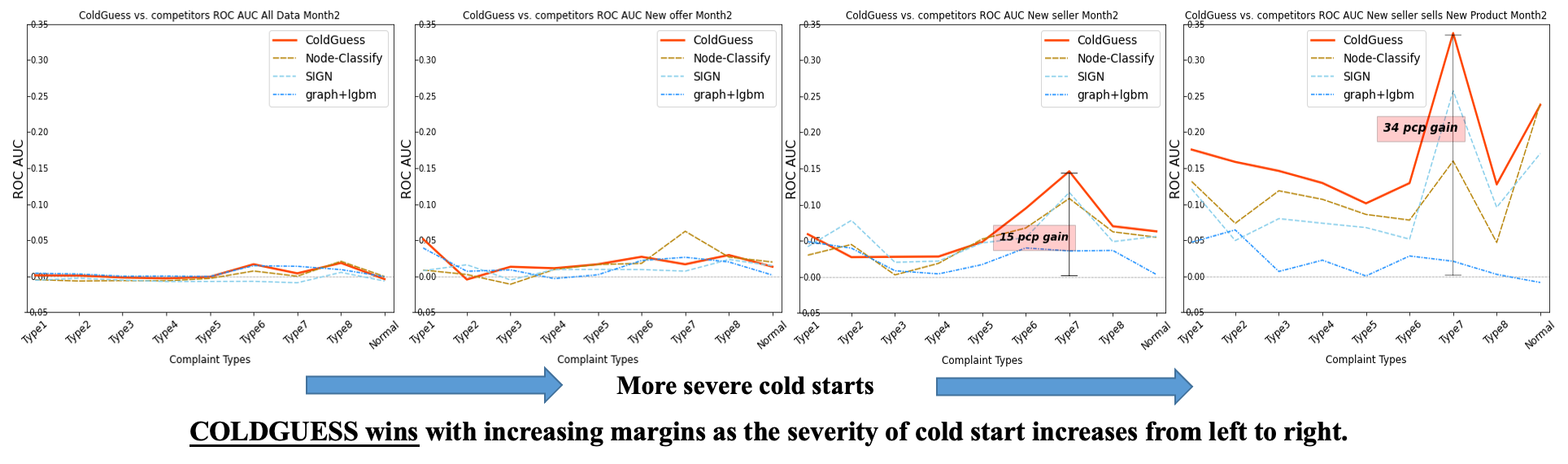}\hfill
    \caption{Performance of \method, \lgbm, \lgbmgraph, SIGN and RGCN on a full spectrum of sample data, new offer case, new seller case, and new seller sell new \ASIN case using data of Month2.}
    \label{fig:202012}
\end{figure*}

\subsection{Data Sets}
We collected 5 seller-\ASIN data sets, including 1 train set and 4 test sets, by taking snapshots of the seller-\ASIN database at the beginning of each month from over 5 consecutive months. Due to confidential limitations, we do not present the detailed statistics of these graphs. 
We only present the two test sets results as Month1 and Month2 in the following sections. The rest of 2 test results are similar and shown in the supplementary. 
We constructed a heterogeneous graph for each month as described in Section~\ref{graph-cons}.

We used customer complaint types as labels. There are eight complaint types based on their root causes, such as expired, defective, damaged, etc. We numbered those complaint types from Type1 to Type8 for illustration purpose. By Adding a Normal type into the classification, we ended up with nine types.

We compared \method with \lgbm, SIGN, RGCN and \lgbmgraph in four scenarios. Below we explained how each scenario is generated to reflect the real world problem. In all scenarios, a complete Seller-\ASINg graph is used for graph model training. 
\begin{itemize}
    \item \textbf{Full spectrum of sample data}: The graph built on the full spectrum data is denoted as $G_o$.
    \item \textbf{New offers case}: We created the new offer set by randomly sampling 25\% of offer listings for each of the 4 minority complaint classes: \USN, \VM, \CCR, and \EPCR, and 1\% of offer listings from the rest of the classes including the Normal one. 
    In this case, new offer edge features are all set as missing except the ones known once an offer is created. %listing price is known as long as the offer is created.
    The resulting graph is denoted as $G_{no}$. 
    \item \textbf{New sellers case}: We created the new seller set by first sampling new offers in the same way as described above. Secondly, we took the sellers associated with the sampled new offers as new sellers. Finally, we extracted all the offers listed under the new sellers as the final new offer set for evaluation. We did not sampled new sellers directly because we want to ensure there is enough data in the minority classes. In this case, the new seller node features and new offer edge features are all set as missing except the ones known once a seller registers or an offer is created. The resulting graph is denoted as $G_{ns}$. 
    % Finally, we removed all the offers listed by certain sellers except the sampled new ones. 
    \item \textbf{New sellers sell new \ASINs case}: We create the new sellers sell new \ASIN case by sampling the new offers and sellers in the same way as described in the \textit{New seller case}. Given the sampled new offers, we took their associated \ASINs as new \ASINs. Then we extracted all the offers listed under either the new sellers or new \ASINs as the final new offer set for evaluation. In this case, new \ASIN and new seller node features, as well as new offer edge features are all set as missing except the ones known once a seller is registered or an offer or \ASIN is created. The resulting graph is denoted as $G_{nsnp}$
    
    %An ASIN's GL group would be known even it is a new ASIN. 
    % Finally, we removed all the offers pointing to certain ASINs except the sampled new ones. 
\end{itemize}
%In the training set, the seller-seller-ASIN graph have 1MM+ ASIN nodes and 255k seller nodes, 1MM+ seller-ASIN edges, 7.4MM phone edges, 287k emails edges, 156k bank account edges, 15k verified phone edges, 7.6k credit card edges, 4.2k international bank account edges, 5.8k cis\_sign\_in edges, 2.6k times\_tin\_id edges.

% \subsubsection{Implementation and Parameter Settings}
% \pcdetector is implemented using DGL~\cite{wang2019deep} and PyTorch~\cite{paszke2019pytorch}. In \method, a node embedder encoded seller and ASIN node features into 64-dim embedding vectors using a 3-layer relational GCN. An edge embedder encoded offer edge features into 105-dim embedding vectors using a 2-layer MLP. A final 2-layer MLP classifier combines the seller, ASIN, offer embedding and predicts the product compliant type. Both \pcdetector and LGBM models were trained on data of 2020/10 and evaluated on the data from 2021/11 to 2021/04. 

\subsection{Performance Evaluation}
In this section we investigated the following two questions:
\begin{itemize}
    \item \textbf{RQ1}: Is \method effective to handle cold start problems?
    \item \textbf{RQ2}: Can \method scale to large graphs with millions of edges?
\end{itemize}

\subsubsection{\textbf{Effectiveness of \method}} \hfill\\
\label{section:effectiveness}
We compared the performance of \method with \lgbm, SIGN, RGCN and \lgbmgraph on all four use cases using $G_o$, $G_{no}$, $G_{ns}$, $G_{nsnp}$ respectively from Month1 and Month2. Figure~\ref{fig:202011} and Figure~\ref{fig:202012} present the results. It can be seen that \method performs fairly well comparing with other methods on a full spectrum of sample data (The left most figures in Figure~\ref{fig:202011} and Figure~\ref{fig:202012}). It starts winning with increasing margins as the severity of cold start cases increases (From the left most figures to right most figures). \method outperforms \lgbm, SIGN, RGCN and \lgbmgraph by up to \textbf{34 \%}, \textbf{11.4 \%}, \textbf{17.3 \%}, \textbf{32.9 \%} respectively. 
Table~\ref{tab:results-month1} and Table~\ref{tab:results-month2} give the detailed performance numbers.

\paragraph{Full spectrum of data} The $G_o$ columns in Table~\ref{tab:results-month1} and Table~\ref{tab:results-month2} show that \method and \lgbmgraph are on par with \lgbm when being evaluated on the entire data set.
%\method occasionally under-performs or outperforms lgbm in 1 or 2 out of 9 classes by less than $0.5\%$. 
SIGN and RGCN slightly under-perform \lgbm because of the nature of GNN. GNNs aggregate neighbors' features to update the target node. When a node feature is a strong indicator, its value will be averaged out by its neighbors. 
The reduced signal makes it more difficult for the model to decide it as risky. \method alleviates this issue by concatenating offer edge features with offer edge's neighboring features through \homoinf. Offer edge features will not be diluted by its neighbors because aggregation and updates only happen on nodes in the \pcdetector's GNN setting.
However, in some cases, \method still performs slightly worse than \lgbm when node features play an important role.
This finding is line with Sergei et al.~\cite{sergei2021boost} that gradient boosted decision trees often outperform other machine learning methods when faced with heterogeneous tabular data. 

% We found RGCN is on the par with lgbm in tested 6 months. It either slightly under-performs or outperforms lgbm for 1 or 2 out of 9 classes by less than 1\% AUC when evaluated on overall data. As we can see in Figure~\ref{fig:origin}, the AUC curve of RGCN overlaps with lgbm most of the time. 
%It is not surprising that RGCN cannot beat lgbm because gradient boosted decision trees (GBDT) often outperform other machine learning methods when faced with heterogeneous tabular data \cite{sergei2021boost}. 

\paragraph{New offer case} 
In this case, new offers' offer features are all set as missing values except the list price. %(104 missing features in total).
We filled those missing values with zeros by default. The $G_{no}$ columns in Table~\ref{tab:results-month1} and Table~\ref{tab:results-month2} show that \method outperforms \lgbm in 8 out of 9 classes by up to 6.5\%.
The reason that \method under-performs \lgbm in \USN is that \USN's top important features are on \ASIN nodes, such as \ASIN age, glance view, etc. As mentioned in the full spectrum of data case, those strong signals are averaged out by their neighbors and thus makes it more difficult for a GNN to learn.
Furthermore, it can be seen that methods leveraging node connectivity information in a graph start outperforming \lgbm in most of the complaint classes.
Meanwhile, \homoinf brings extra gain to \pcdetector. As shown in the tables, \method outperforms \lgbmgraph in 7 out of 9 classes, outperforms SIGN in 8 out of 9 classes and outperform RGCN in 6 out of 9 classes.

\paragraph{New seller case}  
% In this case, we assigned both the offer and seller features to missing values except the ones known once a seller is registered or an offer is created. 
In this case, new sellers and the offers created by these new sellers have their features all set as missing values except the list price. %(160 missing features in total). 
We filled those missing values with zeros by default. The $G_{ns}$ columns in Table~\ref{tab:results-month1} and Table~\ref{tab:results-month2} show that \method outperforms \lgbm in all classes by up to \textbf{12.3\%}. 
It also outperforms \lgbmgraph, SIGN and RGCN in almost all classes by up to \textbf{12.3\%}, 4.3\% and 4.7\% respectively. 
SIGN and RGCN also perform better than \lgbmgraph, which means graph neural networks can improve the performance in cold start cases. 
However, in addition to the benefit of using GNN, \homoinf brings extra performance improvement to \pcdetector.

% compares the performance of \pcdetector and lgbm by using $G_ns$ in 2020/11.
\paragraph{New seller sells a new \ASIN case}  
% Figure~\ref{fig: result}(bottom right) compares the performance of \pcdetector and lgbm by using $G_{nsna}$ in 2020/11. 
In this case, features of new sellers, \ASINs, and offers are all set to missing values except list price and \ASIN product category %(217 missing features in total, and only 2 features left in model). 
The $G_{nsnp}$ columns in Table~\ref{tab:results-month1} and Table~\ref{tab:results-month2} show that \method outperforms \lgbm in all classes by \textbf{9.5-34.1\%}. It also outperforms \lgbmgraph, SIGN and RGCN in all classes by up to \textbf{32.9\%}, \textbf{11.4\%} and \textbf{17.3\%} respectively. Althrough using node connectivity information and GNN improves the model performance of \pcdetector, \homoinf further promotes the model performance by a big margin.

\begin{table}
 \caption{\method wins with increasing margins over \lgbmgraph, SIGN, and RGCN as the severity of cold starts increases. $G_o$, $G_{no}$, $G_{ns}$, $G_{nsnp}$ represent full data, new offer case, new seller case, and new seller sells new \ASIN case in Month1. We take \lgbm as the baseline and present the performance gains of different methods in ROC-AUC.}
%   \caption{\method wins with increasng margin\lgbmgraph, SIGN and RGCN over lgbm on all four use cases using $G_o$, $G_{no}$, $G_{ns}$, $G_{nsnp}$ respectively in Month1. We take the ROC-AUC score of lgbm as the baseline and present the performance improvement of different methods in ROC-AUC using percentage point (pcp).}
\label{tab:results-month1}
\begin{threeparttable}
 \centering
 \begin{tabular}{lrrrr}
  \toprule 
\cmidrule(r){1-4}
\hline
 Defect Type & $G_o$ & $G_{no}$ & $G_{ns}$ & $G_{nsnp}$ \\
  \multicolumn{5}{c}{gains of \method vs. \lgbm} \\
 \hline
\WIS & 0.2\% & \underline{6.4\%} & \underline{6.4\%} & \textbf{\underline{17.8\%}} \\
\USN & -0.2\% & -1.2\% & 2.2\% & \textbf{\underline{16.2\%}} \\
\Damaged & -0.2\% & 1.2\% & 2.3\% & \textbf{\underline{13.8\%}} \\
\OtherNPE & -0.3\% & 1.4\% & 3.1\% & \textbf{\underline{13.5\%}} \\
\Defective & -0.1\% & 1.7\% & 4.5\% & \underline{9.5\%} \\
\VM & 1.5\% & 3.5\% & \textbf{\underline{11.2\%}} & \textbf{\underline{12.7\%}} \\
\CCR & -0.1\% & 1.1\% & \textbf{\underline{12.3\%}} & \textbf{\underline{34.1\%}} \\
\EPCR & 1.2\% & 0.9\% & \underline{5.2\%} & \textbf{\underline{11.9\%}} \\
Normal & -0.4\% & 1.2\% & \underline{5.9\%} & \textbf{\underline{23.2\%}} \\
\hline
 \hline
 \multicolumn{5}{c}{gains of \lgbmgraph vs. \lgbm} \\
 \hline
\WIS & 0.6\% & \underline{5.2\%} & \underline{5.4\%} & \underline{5.2\%} \\
\USN & 0.3\% & 0.7\% & 3.5\% & \underline{6.2\%} \\
\Damaged & 0.0\% & 0.6\% & 0.5\% & 0.5\% \\
\OtherNPE & 0.0\% & -0.1\% & 0.6\% & 2.4\% \\
\Defective & -0.1\% & 0.3\% & 1.7\% & -0.1\% \\
\VM & 1.3\% & 2.4\% & \underline{5.7\%} & 3.2\% \\
\CCR & 1.3\% & 2.8\% & 0.0\% & 1.2\% \\
\EPCR & 0.9\% & 1.4\% & 3.5\% & 0.9\% \\
Normal & -0.1\% & 0.2\% & 0.2\% & -0.7\% \\
\hline
  \multicolumn{5}{c}{gains of SIGN vs. \lgbm }\\
  \hline
\WIS & -0.5\% & 2.5\% & 4.6\% & \textbf{\underline{12.7\%}} \\
\USN & -0.4\% & 1.3\% & \underline{7.3\%} & 4.8\% \\
\Damaged & -0.7\% & -0.7\% & 1.6\% & \underline{6.0\%} \\
\OtherNPE & -0.7\% & 0.8\% & 2.5\% & \underline{7.3\%} \\
\Defective & -0.8\% & -0.2\% & 4.0\% & \underline{5.8\%} \\
\VM & -0.9\% & 0.5\% & \underline{6.9\%} & \underline{6.0\%} \\
\CCR & -0.9\% & 0.5\% & \underline{9.0\%} & \textbf{\underline{26.8\%}} \\
\EPCR & 0.2\% & 0.3\% & 3.2\% & \underline{9.2\%} \\
Normal & -0.6\% & 1.1\% & 4.8\% & \textbf{\underline{15.8\%}} \\
\hline
  \multicolumn{5}{c}{gains of RGCN vs. \lgbm} \\
  \hline
\WIS       & -0.5\% & 2.0\% & 2.8\% & \textbf{\underline{12.6\%}} \\
\USN       & -1.0\% & 0.0\% & 4.1\% & \underline{8.1\%} \\
\Damaged   & -0.7\% & -0.5\% & -0.1\% & \textbf{\underline{11.1\%}} \\
\OtherNPE  & -0.6\% & 0.1\% & 1.9\% & \textbf{\underline{11.0\%}} \\
\Defective & -0.3\% & 0.4\% & 4.6\% & \underline{7.8\%} \\
\VM         & 0.4\% & 2.1\% & \underline{7.9\%} & \underline{9.4\%} \\
\CCR        & -0.1\% & 0.4\% & \underline{7.6\%} & \textbf{\underline{16.8\%}} \\
\EPCR       & 1.1\% & 2.1\% & 4.3\% & 2.9\% \\
Normal      & -0.1\% & 1.8\% & 4.6\% & \textbf{\underline{23.0\%}} \\

\hline
 \end{tabular}
\begin{tablenotes}
      \small
      \item[a] The AUC gain over \lgbm greater than 10 pcp is black bolded and underlined, greater than 5 pcp is underlined. 
\end{tablenotes}
\end{threeparttable}
\end{table}

\begin{table}
 \caption{\method wins with increasing margins over \lgbmgraph, SIGN, and RGCN as the severity of cold starts increases. $G_o$, $G_{no}$, $G_{ns}$, $G_{nsnp}$ represent full data, new offer case, new seller case, and new seller sells new \ASIN case in Month2. We take \lgbm as the baseline and present the performance gains of each model in ROC-AUC.}
%  \caption{The performance improvement of \method, \lgbmgraph, SIGN and RGCN over lgbm on all four use cases using $G_o$, $G_{no}$, $G_{ns}$, $G_{nsnp}$ respectively in Month2. We take the ROC-AUC score of lgbm as the baseline and present the performance improvement of different methods in ROC-AUC using percentage point (pcp).}
\label{tab:results-month2}
\begin{threeparttable}
 \centering
 \begin{tabular}{lrrrr}
  \toprule 
\cmidrule(r){1-4}
\hline
 Defect Type & $G_o$ & $G_{no}$ & $G_{ns}$ & $G_{nsnp}$ \\
 \hline
  \multicolumn{5}{c}{gains of \method vs. \lgbm} \\
 \hline
\WIS & 0.1\% & \underline{5.2\%} & \underline{5.9\%} & \textbf{\underline{17.6\%}} \\
\USN & 0.1\% & -0.5\% & 2.7\% & \textbf{\underline{15.9\%}} \\
\Damaged & -0.2\% & 1.3\% & 2.8\% & \textbf{\underline{14.6\%}} \\
\OtherNPE & -0.3\% & 1.1\% & 2.8\% & \textbf{\underline{12.9\%}} \\
\Defective & -0.1\% & 1.7\% & 4.8\% & \textbf{\underline{10.1\%}} \\
\VM & 1.7\% & 2.7\% & \underline{9.5\%} & \textbf{\underline{12.9\%}} \\
\CCR & 0.4\% & 1.7\% & \textbf{\underline{14.6\%}} & \textbf{\underline{33.7\%}} \\
\EPCR & 1.9\% & 2.9\% &\underline{7.0\%} & \textbf{\underline{12.7\%}} \\
Normal & -0.4\% & 1.3\% & \underline{6.3\%} & \textbf{\underline{23.8\%}} \\
\hline
 \multicolumn{5}{c}{gains of \lgbmgraph vs. \lgbm} \\
 \hline
\WIS & 0.4\% & 3.9\% & 4.9\% & 4.7\% \\
\USN & 0.3\% & 0.7\% & 4.0\% & \underline{6.4\%} \\
\Damaged & 0.0\% & 0.9\% & 0.9\% & 0.7\% \\
\OtherNPE & 0.0\% & -0.3\% & 0.4\% & 2.2\% \\
\Defective & 0.0\% & 0.2\% & 1.7\% & 0.0\% \\
\VM & 1.5\% & 2.2\% & 4.0\% & 2.8\% \\
\CCR & 1.4\% & 2.6\% & 3.6\% & 2.1\% \\
\EPCR & 0.9\% & 2.0\% & 3.7\% & 0.3\% \\
Normal & -0.1\% & 0.2\% & 0.3\% & -0.9\% \\
\hline
  \multicolumn{5}{c}{gains of SIGN vs. \lgbm} \\
  \hline
\WIS & -0.6\% & 0.8\% & 4.2\% & \textbf{\underline{12.1\%}} \\
\USN & -0.3\% & 1.6\% & \underline{7.8\%} & 5.0\% \\
\Damaged & -0.5\% & -0.5\% & 2.0\% & \underline{8.0\%} \\
\OtherNPE & -0.7\% & 0.9\% & 2.2\% & \underline{7.4\%} \\
\Defective & -0.7\% & 0.9\% & 4.7\% & \underline{6.8\%} \\
\VM & -0.7\% & 0.9\% & \underline{5.4\%} & \underline{5.2\%} \\
\CCR & -0.9\% & 0.7\% & \textbf{\underline{11.7\%}} & \textbf{\underline{25.7\%}} \\
\EPCR & 0.5\% & 2.3\% & 4.87\% & \underline{9.6\%} \\
Normal & -0.7\% & 1.5\% & \underline{5.6\%} & \textbf{\underline{17.0\%}} \\

\hline
  \multicolumn{5}{c}{gains of RGCN vs. \lgbm} \\
  \hline
\WIS       & -0.5\% & 0.9\% & 3.0\% & \textbf{\underline{13.2\%}} \\
\USN       & -0.7\% & 0.2\% & 4.5\% & \underline{7.4\%} \\
\Damaged   & -0.6\% & -1.1\% & 0.3\% & \textbf{\underline{11.9\%}} \\
\OtherNPE  & -0.6\% & 0.9\% & 1.9\% & \textbf{\underline{10.7\%}} \\
\Defective & -0.3\% & 1.6\% & \underline{5.2\%} & \underline{8.6\%} \\
\VM        & 0.7\% & 1.8\% & \underline{6.8\%} & \underline{7.8\%} \\
\CCR       & 0.0\% & \underline{6.2\%} & \textbf{\underline{10.9\%}} & \textbf{\underline{16.0\%}} \\
\EPCR      & 2.1\% & 2.7\% & \underline{6.2\%} & 4.7\% \\
Normal     & -0.1\% & 2.0\% & \underline{5.5\%} & \textbf{\underline{23.98\%}} \\

\hline
 \end{tabular}
\begin{tablenotes}
      \small
      \item[a] The AUC gain over \lgbm greater than 10 pcp is black bolded and underlined, greater than 5 pcp is underlined. 
\end{tablenotes}
\end{threeparttable}
\end{table}

%Table~\ref{tab:results-month1} and Table~\ref{tab:results-month2} presents the detailed performance of \lgbmgraph and \pcdetector over lgbm on two different months' data. It can be seen that \pcdetector consistently outperforms lgbm and \lgbmgraph by \textbf{9.5-34.1\%} and \textbf{9.5-32.9\%} absolute AUC respectively in the new seller sells new \ASIN case. It can also outperforms lgbm by \textbf{2.2-14.6\%} absolute AUC in the new seller case and outperforms \lgbmgraph by upto \textbf{12.3\%} absolute AUC in most cases of the new seller case.
%The significant AUC gain in \method over \lgbmgraph suggests that filling seller missing features is not enough when most of offer and \ASIN features are missing. However, we found the performance of \lgbmgraph does not lift much after filling offer's missing features using two-hops graph propagation. It indicates that filling missing features with meaningful information not only relies on the graph structure, but also relies on the weights assigned on different relation types as well as the multi-hops propagation, which allows the risk passing across heterogeneous nodes. 

In summary, though \pcdetector, \lgbm, \lgbmgraph, SIGN, RGCN are tied when being evaluated on the full spectrum of sample data, \method outperforms all competitors in three cold start cases on 4 different datasets. The stable superb performance of \method, especially in the \textit{New seller} and \textit{New seller sells a new \ASIN} cases, comes from its advantage of leveraging node connectivity information, graph neural network and \homoinf. 
\method outperforming SIGN and RGCN shows that \textit{\homoinf} can further boost the model performance even when node connectivity information and graph neural network are already used.

%\method outperforms \lgbmgraph and \lgbmgraph outperforms lgbm in three cold start cases over six data sets for each case. The stable superb performance of \pcdetector, especially in the \textit{New seller} and \textit{New seller sells a new \ASIN} cases, mainly comes from its advantage of handling missing features with linkage information. The linkage information has proven to be valuable, especially in cold start cases. We found that on average 43\% of new sellers are connected with existing sellers via various relations, and a majority of sellers who connected to risky sellers are also found risky. Given the connectivity information embedded in graph, \method fills the missing node features by aggregating information from neighboring nodes. For missing edge features, which cannot be handled by classic RGCN, \method enriches the edge features by concatenating it and its neighbors’. In this way, \pcdetector leverages graph connections to improve the prediction accuracy through multi-layer message passing. As a contrast, lgbm is not able to utilize those information.

%\lgbmgraph also leverages the seller-seller relations to fill missing seller features. It explains its superior performance over lgbm in all three cold start cases. However, \lgbmgraph under-performs \method. It suggests that assigning optimized weights on relation types and the multi-hops propagation process also improves prediction. 

\begin{figure}
\centering
\begin{subfigure}{.2\textwidth}
  \centering
  \includegraphics[width=1.2\linewidth]{Figures/train_time.png}
  \caption{Training time}
  \label{fig_train}
\end{subfigure}
\quad
%   \hskip -1ex
\begin{subfigure}{.2\textwidth}
  \centering
  \includegraphics[width=1.2\linewidth]{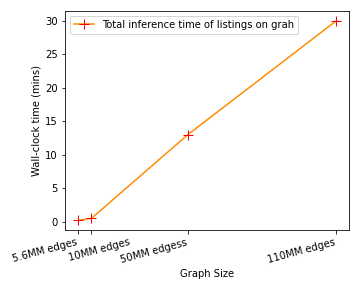}
  \caption{Inference time}
  \label{fig_inference}
\end{subfigure}
\caption{\method scales linearly: training time vs. graph size (left), and inference time vs. graph size (right)}
\label{fig:scalability}
\end{figure}

\subsubsection{\textbf{Scalability of \method}} \hfill\\
We studied the training and inference scalability of \method by using graphs with different sizes. All experiments are conducted on a p3.8xlarge EC2 instance equipped with 4 V100 GPU.
During model training, \method went over all labelled offer listings in each epoch.\footnote{The number of labelled data increases linearly with the number of edges in a graph.} During model inference, \method went over all the offer listings as we do in the product system.
Figure~\ref{fig_train} presents the training time of \method on graphs scaling from 2.5M edges to 10M edges. It can be seen that the training time of \method scales linearly with the number of edges in a graph. 
Figure~\ref{fig_inference} presents the inference time of \method on graphs scaling from 5.6M edges to 110M edges. It can be seen that the inference time of \method scales linearly with the number of edges in a graph.

%Among all the complaint types, with the increasing number of missing features, the performance drop of CCR predicted by lgbm is most significant. The performance drop of EPCR is the least because only a few GL product groups contains perishable ASINs \footnote{Food can be expired but GL such as electronics and toys will not have expired items}. As a strong indicator of expiration, GL product group uphold the lgbm performance better than other defect classes. 

\subsection{Ablation Study}
Table~\ref{tab:ablation} demonstrates the efficacy of each component of \method. For easy comparison, we summarize each model's performance by aggregating the ROCAUC across nine classes using geometric mean\footnote{We tried both geometric mean and harmonic mean for aggregation, and found they gave similar results}. Table~\ref{tab:ablation} compares model performance relative to lgbm in four evaluation cases. 

Node connectivity plus lgbm (i.e. \lgbmgraph) outperforms lgbm by +0.5\% to +2.4\% AUC. Node connectivity plus relation type information (i.e. RGCN) improves the performance of \lgbmgraph in the last two severe cold start cases by +2.5\% to 8.1\% AUC\footnote{In non-cold start cases, we observe lgbm works better than graph-based models as later will average out significant signals in neighbor aggregation. See \ref{section:effectiveness} for details}. Consolidating nodes in RGCN plus \homoinfps (i.e. \method) further improves performance of RGCN by +0.4\% to 5.6\% AUC. The design decision of consolidating nodes and using \homoinfps has the highest impact on the prediction accuracy(up to 5.6\% AUC gain).

% While \lgbmgraph uses equal relation weight as 1 and then lgbm for classification, SIGN uses Laplacian transformation of node degrees to calculate the relation weights and then MLP for classification. SIGN works better than \lgbmgraph in the last two severe cold start cases by +2.5\% to 8.1\% AUC, suggesting that assigning different weights on relation types helps with the increased severity of cold start, and lgbm works better in non-cold start cases. The differences between RGCN and SIGN are (1) RGCN learns the relation weight from the optimization process (2) Like \method, SIGN generates seller and \ASINs embeddings through graph propagation and then concatenates them with offer edge embeddings. As contrast, RGCN uses offer as node and propagate offer features from offer to seller to \ASIN and then to another offer. \method simplified RGCN using consolidated nodes and \homoinf, meaning offer are used as edges, and offer edges are not involved into propagation so that the influence is direct from offer to offer. RGCN outperforms SIGN in 3 out of 4 cases, suggesting that the relation weights learned by model works better than Laplacian transformation. \method outperforms both SIGN and RGCN by +0.4\% to +5.6\% AUC, demonstrating both \homoinf and \consolidated further improves the prediction. 

\begin{table}
 \caption{Ablation study shows the efficacy of each component of \method. We report model performance relative to lgbm using ROCAUC.}
\label{tab:ablation}
\begin{threeparttable}
 \centering
 \begin{tabular}{l|rrrr}
  \toprule 
\cmidrule(r){1-4}
\hline
Method & $G_o$ & $G_{no}$ & $G_{ns}$ & $G_{nsnp}$ \\
 \hline
%\parbox[t]{2mm}{\multirow{5}{*}{\rotatebox[origin=c]{90}{\textit{Ablation}}}}
lgbm & 0.0\% & 0.0\% & 0.0\% & 0.0\% \\
\lgbmgraph & 0.5\% & 1.5\% & 2.4\% & 2.3\% \\
%SIGN & -0.6\% & 0.7\% & 4.9\% & 10.4\% \\
RGCN & -0.2\% & 0.9\% & 4.3\% & 11.6\% \\
\method & 0.2\% & 1.8\% & 5.9\%  & 17.2\% \\
\hline
 \end{tabular}
% \begin{tablenotes}
    %   \small
    %   \item[a] The AUC gain over \lgbm greater than 10 pcp is black bolded and underlined, greater than 5 pcp is underlined. 
% \end{tablenotes}
\end{threeparttable}
\end{table}

\section{Conclusion and Future Work}
In this paper, we proposed \pcdetector, a product complaint detector which deals with highly multi-relational large-scale graph and addresses the cold start problems via \textit{\consolidated} and \textit{\homoinf}. \method enjoys the following advantages over traditional models:
\begin{itemize}
    \item \textbf{General:} This inductive framework can handle dynamic heterogeneous graphs with tens of millions of nodes, hundreds of millions of edges, as well as massive node and edge features.
    \item \textbf{Effective}, especially for \textit{Cold Start Problems}. It outperforms \lgbm in all three cold start scenarios by \textit{up to 34 pcp} AUC, and demonstrates stable performance with increased number of missing values.
    \item \textbf{Fast and Scalable:} \method scales linearly with the input size, as shown in ~\ref{fig:scalability}. It makes 1MM+ predictions in less than 2 mins on a 4-GPU P3.8xlarge EC2 instance.
\end{itemize}
\method has superior performance in cold start problems and is already in production. Our future work includes adding temporal information so that we can combine multiple months of data to augment the data size of rare classes in training.

% \method, this large-scale graph based product complaint detector will \textit{be in production} in early July of 2021, making daily predictions on 55MM+ listings with a goal of 5\% YoY reduction in customer complaints.

\bibliographystyle{ACM-Reference-Format}
\bibliography{reference.bib}
\citestyle{acmauthoryear}

\pagebreak
\clearpage
\section{Supplementary Material}
Table~\ref{tab:results-month3} and Table~\ref{tab:results-month4} shows that \method wins with increasing margins over lgbm, \lgbmgraph, SIGN, and RGCN as the severity of cold starts increases on Month3 and Month4.

\begin{table}[b]
 \caption{\method wins with increasing margins over \lgbmgraph, SIGN, and RGCN as the severity of cold starts increases. $G_o$, $G_{no}$, $G_{ns}$, $G_{nsnp}$ represent full data, new offer case, new seller case, and new seller sells new \ASIN case in Month3. We take lgbm as the baseline and present the performance gains of different methods in ROC-AUC.}
\label{tab:results-month3}
\begin{threeparttable}
 \centering
 \begin{tabular}{lrrrr}
  \toprule 
\cmidrule(r){1-4}
\hline
 Defect Type & $G_o$ & $G_{no}$ & $G_{ns}$ & $G_{nsnp}$ \\
  \multicolumn{5}{c}{gains of \method vs. lgbm} \\
 \hline
\WIS  &  0.2\%  &  \underline{5.5\%}  &  \underline{5.2\%}  &  \textbf{\underline{17.3\%}}  \\
\USN  &  0.3\%  &  -0.4\%  &  3.1\%  &  \textbf{\underline{16.6\%}}  \\
\Damaged  &  -0.2\%  &  1.1\%  &  2.7\%  &  \textbf{\underline{14.6\%}}  \\
\OtherNPE  &  -0.2\%  &  1.1\%  &  2.6\%  &  \textbf{\underline{12.8\%}}  \\
\Defective  &  -0.1\%  &  1.6\%  &  4.9\%  &  \underline{9.8\%}  \\
\VM &  1.0\%  &  1.7\%  &  \underline{8.0\%}  &  \textbf{\underline{14.5\%}}  \\
\CCR  &  -0.6\%  &  1.9\%  &  \textbf{\underline{12.4\%}}  &  \textbf{\underline{33.0\%}}  \\
\EPCR  &  3.1\%  &  4.3\%  &  \textbf{\underline{12.4\%}}  &  \textbf{\underline{15.0\%}}  \\
Normal  &  -0.2\%  &  1.5\%  &  \underline{6.3\%}  &  \textbf{\underline{22.9\%}}  \\
\hline
 \hline
 \multicolumn{5}{c}{gains of \lgbmgraph vs. lgbm} \\
 \hline
\WIS  &  0.7\%  &  4.4\%  &  4.5\%  &  \underline{5.1\%}  \\
\USN  &  0.4\%  &  0.8\%  &  4.0\%  &  \underline{6.4\%}  \\
\Damaged  &  0.1\%  &  0.9\%  &  0.8\%  &  0.6\%  \\
\OtherNPE  &  0.1\%  &  -0.4\%  &  0.4\%  &  2.6\%  \\
\Defective  &  0.0\%  &  0.4\%  &  2.0\%  &  -0.1\%  \\
\VM &  1.4\%  &  1.7\%  &  2.5\%  &  3.8\%  \\
\CCR  &  2.1\%  &  4.1\%  &  4.6\%  &  1.4\%  \\
\EPCR  &  2.5\%  &  3.4\%  &  \underline{8.3\%}  &  0.4\%  \\
Normal  &  0.1\%  &  0.4\%  &  -0.2\%  &  -0.9\%  \\
\hline
  \multicolumn{5}{c}{gains of SIGN vs. lgbm }\\
  \hline
\WIS  &  -0.2\%  &  1.8\%  &  3.6\%  &  \textbf{\underline{11.3\%}}  \\
\USN  &  -0.2\%  &  1.9\%  &  \underline{7.9\%}  &  \underline{5.7\%}  \\
\Damaged  &  -0.6\%  &  -0.8\%  &  1.9\%  &  \underline{7.1\%}  \\
\OtherNPE  &  -0.8\%  &  0.5\%  &  1.4\%  &  \underline{7.7\%}  \\
\Defective  &  -0.6\%  &  1.4\%  &  \underline{5.0\%}  &  \underline{7.6\%}  \\
\VM &  -3.5\%  &  -3.0\%  &  2.9\%  &  \underline{8.0\%}  \\
\CCR  &  -0.7\%  &  1.3\%  &  \underline{9.4\%}  &  \textbf{\underline{25.7\%}}  \\
\EPCR  &  0.9\%  &  2.2\%  &  \underline{7.2\%}  &  \textbf{\underline{11.0\%}}  \\
Normal  &  -0.4\%  &  1.5\%  &  \underline{6.4\%}  &  \textbf{\underline{16.1\%}}  \\
\hline
  \multicolumn{5}{c}{gains of RGCN vs. lgbm} \\
  \hline
\WIS  &  -0.3\%  &  1.8\%  &  2.1\%  &  \textbf{\underline{12.1\%}}  \\
\USN  &  -0.5\%  &  0.9\%  &  4.5\%  &  \underline{7.7\%}  \\
\Damaged  &  -0.6\%  &  -1.4\%  &  0.1\%  &  \textbf{\underline{11.9\%}}  \\
\OtherNPE  &  -0.5\%  &  0.4\%  &  1.7\%  &  \textbf{\underline{10.6\%}}  \\
\Defective  &  -0.4\%  &  1.7\%  &  \underline{5.3\%}  &  \underline{8.1\%}  \\
\VM &  0.4\%  &  1.4\%  &  3.1\%  &  \underline{7.5\%}  \\
\CCR  &  0.0\%  &  1.1\%  &  \underline{8.7\%}  &  \textbf{\underline{15.9\%}}  \\
\EPCR  &  3.0\%  &  4.7\%  &  \textbf{\underline{11.5\%}}  &  \underline{6.9\%}  \\
Normal  &  0.0\%  &  2.0\%  &  \underline{6.4\%}  &  \textbf{\underline{24.1\%}}  \\
\hline
 \end{tabular}
\begin{tablenotes}
      \small
      \item[a] The AUC gain over lgbm greater than 10 pcp is black bolded and underlined, greater than 5 pcp is underlined. 
\end{tablenotes}
\end{threeparttable}
\end{table}

\begin{table}[b]
 \caption{\method wins with increasing margins over \lgbmgraph, SIGN, and RGCN as the severity of cold starts increases. $G_o$, $G_{no}$, $G_{ns}$, $G_{nsnp}$ represent full data, new offer case, new seller case, and new seller sells new \ASIN case in Month4. We take lgbm as the baseline and present the performance gains of different methods in ROC-AUC.}
\label{tab:results-month4}
\begin{threeparttable}
 \centering
 \begin{tabular}{lrrrr}
  \toprule 
\cmidrule(r){1-4}
\hline
 Defect Type & $G_o$ & $G_{no}$ & $G_{ns}$ & $G_{nsnp}$ \\
  \multicolumn{5}{c}{gains of \method vs. lgbm} \\
 \hline
\WIS  &  0.2\%  &  \underline{5.8\%}  &  \underline{6.2\%}  &  \textbf{\underline{17.5\%}}  \\
\USN  &  0.1\%  &  0.1\%  &  2.9\%  &  \textbf{\underline{16.5\%}}  \\
\Damaged  &  -0.2\%  &  1.8\%  &  3.1\%  &  \textbf{\underline{15.5\%}}  \\
\OtherNPE  &  -0.1\%  &  1.7\%  &  3.5\%  &  \textbf{\underline{12.5\%}}  \\
\Defective  &  -0.1\%  &  2.0\%  &  \underline{5.5\%}  &  \textbf{\underline{10.0\%}}  \\
\VM &  0.9\%  &  2.2\%  &  \underline{7.1\%}  &  \textbf{\underline{15.1\%}}  \\
\CCR  &  0.5\%  &  2.0\%  &  \textbf{\underline{15.1\%}}  &  \textbf{\underline{36.0\%}}  \\
\EPCR  &  3.8\%  &  \underline{5.1\%}  &  \textbf{\underline{12.9\%}}  &  \textbf{\underline{16.7\%}}  \\
Normal  &  -0.4\%  &  1.6\%  &  \underline{7.4\%}  &  \textbf{\underline{25.4\%}}  \\
\hline
 \hline
 \multicolumn{5}{c}{gains of \lgbmgraph vs. lgbm} \\
 \hline
\WIS  &  0.6\%  &  4.4\%  &  \underline{5.0\%}  &  4.5\%  \\
\USN  &  0.5\%  &  1.1\%  &  4.1\%  &  \underline{6.2\%}  \\
\Damaged  &  0.0\%  &  1.3\%  &  1.0\%  &  0.6\%  \\
\OtherNPE  &  0.1\%  &  0.0\%  &  0.8\%  &  2.2\%  \\
\Defective  &  0.0\%  &  0.5\%  &  2.3\%  &  0.1\%  \\
\VM &  1.4\%  &  2.2\%  &  2.4\%  &  3.4\%  \\
\CCR  &  2.8\%  &  3.7\%  &  \underline{7.3\%}  &  3.2\%  \\
\EPCR  &  2.7\%  &  3.4\%  &  \underline{6.3\%}  &  1.4\%  \\
Normal  &  0.1\%  &  0.4\%  &  0.2\%  &  -0.8\%  \\
\hline
  \multicolumn{5}{c}{gains of SIGN vs. lgbm }\\
  \hline
\WIS  &  -0.3\%  &  2.1\%  &  4.7\%  &  \textbf{\underline{11.4\%}}  \\
\USN  &  0.0\%  &  3.1\%  &  \underline{8.3\%}  &  \underline{6.1\%}  \\
\Damaged  &  -0.6\%  &  -0.4\%  &  2.5\%  &  \underline{8.6\%}  \\
\OtherNPE  &  -0.8\%  &  1.6\%  &  2.4\%  &  \underline{7.9\%}  \\
\Defective  &  -0.5\%  &  1.7\%  &  \underline{5.7\%}  &  \underline{7.4\%}  \\
\VM &  -4.3\%  &  -3.3\%  &  2.2\%  &  \textbf{\underline{10.0\%}}  \\
\CCR  &  0.1\%  &  1.1\%  &  \textbf{\underline{10.9\%}}  &  \textbf{\underline{28.1\%}}  \\
\EPCR  &  1.3\%  &  3.2\%  &  \underline{7.4\%}  &  \textbf{\underline{12.6\%}}  \\
Normal  &  -0.6\%  &  1.6\%  &  \underline{7.1\%}  &  \textbf{\underline{17.9\%}}  \\
\hline
  \multicolumn{5}{c}{gains of RGCN vs. lgbm} \\
  \hline
\WIS  &  -0.3\%  &  2.4\%  &  3.4\%  &  \textbf{\underline{12.9\%}}  \\
\USN  &  -0.5\%  &  1.6\%  &  4.6\%  &  \underline{8.2\%}  \\
\Damaged  &  -0.5\%  &  -0.1\%  &  1.0\%  &  \textbf{\underline{13.2\%}}  \\
\OtherNPE  &  -0.3\%  &  1.7\%  &  3.0\%  &  \textbf{\underline{10.8\%}}  \\
\Defective  &  -0.3\%  &  1.8\%  &  \underline{6.0\%}  &  \underline{8.7\%}  \\
\VM &  -0.3\%  &  5.7\%  &  4.8\%  &  \underline{9.5\%}  \\
\CCR  &  0.9\%  &  1.5\%  &  \textbf{\underline{10.6\%}}  &  \textbf{\underline{17.8\%}}  \\
\EPCR  &  3.5\%  &  4.5\%  &  \textbf{\underline{11.0\%}}  &  \underline{7.8\%}  \\
Normal  &  0.0\%  &  1.6\%  &  \underline{6.9\%}  &  \textbf{\underline{25.9\%}}  \\
\hline
 \end{tabular}
\begin{tablenotes}
      \small
      \item[a] The AUC gain over lgbm greater than 10 pcp is black bolded and underlined, greater than 5 pcp is underlined. 
\end{tablenotes}
\end{threeparttable}
\end{table}

\end{document}